\documentclass[sigconf]{acmart}
\usepackage{algorithm}
\usepackage{algpseudocode}
\usepackage{graphicx}
\usepackage{multicol}
\usepackage{amsmath}

\usepackage{rotating}
\usepackage{amsthm}
\usepackage{soul}
\usepackage{mathrsfs}
\usepackage{pgfplots}
\pgfplotsset{compat=1.18}
\usepackage{pgfplots}
\usepackage{svg}
\usepackage{adjustbox}
\usepackage{subcaption}

\usepackage{booktabs}
\usepackage{adjustbox}
\usepackage{graphicx}
\usepackage{bm} 
\usepackage{caption} 
\usepackage{bbding}
\DeclareCaptionFormat{myformat}{\textbf{#1}#2#3}
\AtBeginDocument{%
  \providecommand\BibTeX{{%
    \normalfont B\kern-0.5em{\scshape i\kern-0.25em b}\kern-0.8em\TeX}}}

\copyrightyear{2023}
\acmYear{2023}
\setcopyright{acmlicensed}\acmConference[CIKM '23]{Proceedings of the 32nd
ACM International Conference on Information and Knowledge
Management}{October 21--25, 2023}{Birmingham, United Kingdom}
\acmBooktitle{Proceedings of the 32nd ACM International Conference on
Information and Knowledge Management (CIKM '23), October 21--25, 2023,
Birmingham, United Kingdom}
\acmPrice{15.00}
\acmDOI{10.1145/3583780.3615273}
\acmISBN{979-8-4007-0124-5/23/10}

\begin{document}


\title{Leveraging Knowledge and Reinforcement Learning for Enhanced Reliability of Language Models}

\settopmatter{printfolios=true}


\author{
    \texorpdfstring{Nancy Tyagi, Surjodeep Sarkar, Manas Gaur}{Nancy Tyagi, Surjodeep Sarkar, Manas Gaur} \\
    University of Maryland, Baltimore County, MD, United States \\
    \{nancyt1, ssarkar1, manas\}@umbc.edu
}

\renewcommand{\shortauthors}{Tyagi, et al.}

\begin{abstract}
The Natural Language Processing (NLP) community has been using crowd-sourcing techniques to create benchmark datasets such as General Language Understanding and Evaluation (GLUE) for training modern Language Models (LMs) such as BERT. GLUE tasks measure the reliability scores using inter-annotator metrics - Cohen's Kappa ($\kappa$). However, the reliability aspect of LMs has often been overlooked. To counter this problem, we explore a knowledge-guided LM ensembling approach that leverages reinforcement learning to integrate knowledge from ConceptNet and Wikipedia as knowledge graph embeddings. This approach mimics human annotators resorting to external knowledge to compensate for information deficits in the datasets. Across nine GLUE datasets, our research shows that ensembling strengthens reliability and accuracy scores, outperforming state-of-the-art. 
\end{abstract}




\maketitle

\section{Introduction}
The NLP community is growing by developing new LMs and datasets catering to a wide range of domains, including general-purpose \cite{wang2019superglue} and domain-specific \cite{xiao-etal-2021-ernie}. Concurrently, there is an emergent unease about the ability of these new LMs to emulate the performance derived from human annotations in such datasets, which is typically assessed via inter-annotator agreement scores, for instance, Cohen's Kappa ($\kappa$) \cite{mchugh2012interrater}.
However, performance assessment predominantly hinges on conventional metrics, which do not adequately reflect the reliability of LMs \cite{liang2022holistic}. Nonetheless, every new LMs to achieve acceptance within the NLP community has to demonstrate effectiveness in understanding natural language through simple and effective GLUE benchmarks \cite{wang2018glue}. GLUE benchmarks have established prominence in NLP because of high annotator agreement, thus defining a high threshold for new LMs to break.  Interestingly, since the inception of the GLUE benchmarks, no prior work has emphasized the use of annotation agreement as a proxy measure for reliability in LMs. As a countermeasure, researchers have been increasingly allocating resources to advance LMs, but this approach has inadvertently compromised the model's ability to surpass simpler LMs or human performance, especially when the models are trained on datasets aggregated via crowdsourcing. We present an ensembling of LMs, taking inspiration from \textit{Cohen's Kappa}, which states that if an annotation agrees with two annotators, it is sufficiently reliable \cite{mchugh2012interrater}. Ensembling of LMs presents a synergistic collaboration among simpler models, culminating into a system that is more resilient and effective than singular models. We emphasize that the ensemble's collective strength enables it to compensate for the inadequacies of an individual model under specific conditions and bolster its decision-making confidence. Their performance needs to be assessed for the ensembling of LMs to be functional and reliable.

In our study, we aim to conceptualize, devise, and evaluate the ensembling of LMs by addressing three research questions:
\textbf{RQ1:} Can we employ $\kappa$ to evaluate the reliability of LMs trained on GLUE benchmarks?
\textbf{RQ2:} Considering the language models as annotators, is it possible to enhance $\kappa$ by strategically ensembling LMs?
\textbf{RQ3:} Given that crowd workers frequently resort to external knowledge to augment the quality of annotations, can the infusion of external knowledge during ensembling improve overall reliability?
To answer these questions, we make two contributions : (a) We propose three ensembling techniques: Shallow Ensemble (ShE), Semi Ensemble (SE), and Deep Ensemble (DE), where DE is characterized as a knowledge-guided ensembling method that integrates LMs with knowledge from ConceptNet \cite{speer2017conceptnet} and Wikipedia \cite{yamada2020wikipedia2vec} through reinforcement learning (RL) \cite{rawte2022tdlr}. (b) We evaluate the reliability of the ensemble models using $\kappa$ across nine GLUE tasks. The paper is structured as follows: Section 2 introduces the three Ensemble methods. Section 3 covers experimental details, including datasets, metrics, and models used. Section 4 discusses the results, and Section 5 concludes the paper.

\begin{figure*}[t]
 \centering
 \adjustbox{trim=0cm 4cm 0cm 0cm}{%
   \includegraphics[width=\textwidth]{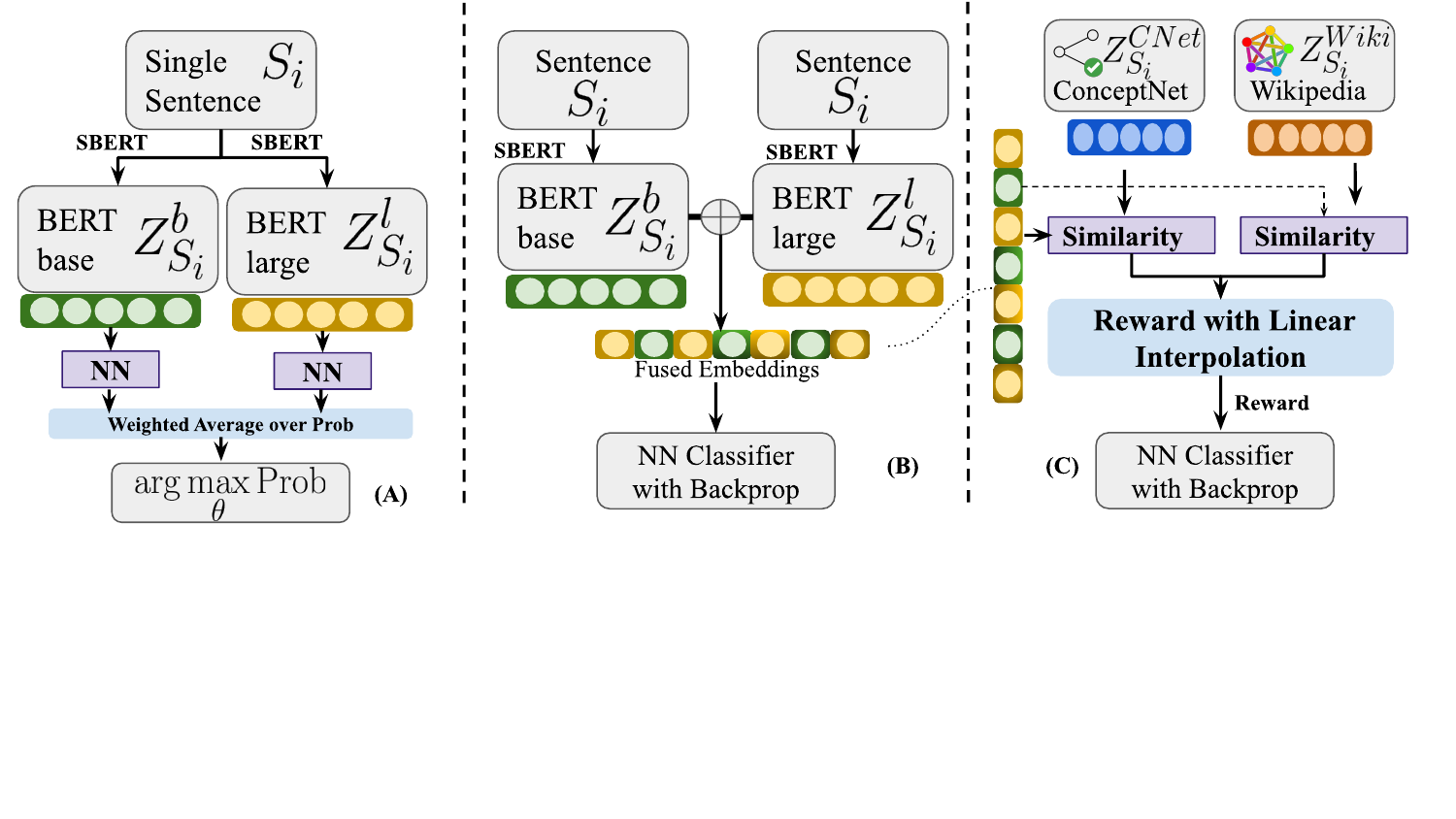}
}
 \caption{Illustration of the three proposed Ensemble Methods on two variants of the BERT model: (A) A weighted average driven Shallow-Ensemble (ShE) method, (B) an embedding fused Semi-Ensemble (SE) method, and (C) A knowledge guided Deep-Ensemble (DE) method which uses external knowledge from Wikipedia and ConceptNet knowledge graph.}
\label{fig:fig1}
\end{figure*}


\section{ENSEMBLE METHODS}
Let  $\mathcal{D}=\{S^{(i)}, y^{(i)}: i=1, \dots, m\}$ be the given dataset, where $S^{(i)}$ is the text sentence and $y^{(i)}$ is the observed class for the $i$th sentence. $y^{(i)}$ can take values from 1 to $c$. Let $\mathcal{M}=\{M_{\ell}: \ell=1, \dots, n\}$ be a collection of $n$~ LMs. $S^{(i)}$ is transformed to a feature vector as 
\begin{equation}\label{eq:embeddings}
Z^{M_l}_{S_i} = \text{SBERT}(S_i, M_l)
\end{equation} 
, where SBERT represents sentence transformer \cite{reimers2019sentence}. We take the embeddings generated using SBERT because it outperforms individual BERT embeddings \cite{devlin2018bert}. 

\textbf{1. Shallow-Ensemble (ShE):} For each $Z^{M_l}_{S_i}$, the estimated probability of it belonging to a certain category $k$ using model $M_{\ell}$ is denoted as $\text{Prob}_\ell(y_i = k | Z^{M_l}_{S_i})$. Given weights $\alpha_1, \ldots, \alpha_n$ such that $\alpha_{\ell} \in [0, 1]$ and $\sum_{\ell=1}^n \alpha_{\ell} = 1$, the probabilities are combined as 
$
\sum_{\ell=1}^{n} \alpha_\ell \cdot \text{Prob}_\ell(y_i = k | x_i) 
$ as shown in Figure \ref{fig:fig1} (A).
The predicted class is obtained as
\begin{equation}
\hat{y}_i(\alpha) = \arg \max_k \left[\sum_{\ell=1}^{n} \alpha_\ell \cdot \text{Prob}_\ell(y(i) = k | Z^{M_l}_{S_i})\right].
\label{eq:she1}
\end{equation} 
The loss is defined as a function of $\alpha$ as
$
L(\alpha) = \sum_{i=1}^{m} \mathbb{I}[y_i \neq \hat{y}_i(\alpha)]
$.
The objective is to minimize the loss for better performance. ShE uses a statistical approach of averaging the predicted probabilities. 

\textbf{2. Semi-Ensemble (SE):} We define a new feature vector $Z^{\prime M_l}_{S_i}$ as 
$
Z^{\prime M_l}_{S_i} = Z^{M_1}_{S_i} \oplus Z^{M_2}_{S_i} \oplus \cdot \cdot \cdot \oplus Z^{M_l}_{S_i}
$, which represents the fused embeddings obtained from $M_l$ models as described in Figure \ref{fig:fig1} (B). These fused embeddings leverage the combined strength of individual models. These embeddings are then fed into a Neural Network (NN). The objective is to minimize the Binary Cross Entropy loss function defined by L as
\[
L = -\frac{1}{N} \sum_{i=1}^{N} y_i \log(\hat{y}_i) + (1-y_i) \log(1-\hat{y}_i)
\]

where $N$ is the total number of samples, $y_i$ represents the ground truth label for the $i$-th sample, and $\hat{y}_i$ represents the predicted probability output by the NN for the $i^{th}$ sample. 

\textbf{3. Deep-Ensemble (DE):} We incorporate external knowledge using two general-purpose knowledge graphs: ConceptNet(CNet) and Wikipedia(Wiki) (as shown in \ref{fig:fig1}(C)), to improve SE ensemble model. This addition helps in contextual understanding of LMs. For each $S_{i}$, we denote the 
embeddings from CNet and Wiki as $Z^{CNet}_{S_i}$ and $Z^{Wiki}_{S_i}$ respectively.Using the fused embeddings $Z^{\prime M_l}_{S_i}$, the reward of the RL policy is computed as: 
\begin{equation}
\begin{split}
R({\beta})_{i} := \beta \, \text{CS}(Z^{\text{CNet}}_{S_i},Z^{\prime M_l}_{S_i})
+ (1-\beta) \, \text{CS}(Z^{Wiki}_{S_i},Z^{\prime M_l}_{S_i})
\end{split}
\label{eq:sim_metric}
\end{equation}
 where $\beta_{i} \in [0, 1]$ for $i = 1, 2, \ldots, n$, $\sum_{\ell=1}^n \beta_{\ell} = 1$, and $CS$ denotes cosine similarity. The loss is defined as a function of ${\beta}$ as 
\begin{equation}
    L(\beta) = \frac{1}{N} \sum_{i=1}^{N} \bigg(y_i \log(\hat{y}_i) + (1-y_i) \log(1-\hat{y}_i)\bigg) \cdot R(\beta)_{i}
\end{equation}
where $N$ is the total number of samples, $y_i$ is the ground truth label and $\hat{y}_i$ represents the predicted probability output by the NN for the $i^{th}$ sample. The objective is to minimize the loss by maximizing $R({\beta})$. Table \ref{table:functionality} describes the different components of the three ensemble methods.

\begin{table}[t]
    \centering
    \begin{tabular}{p{3.5cm}|p{0.5cm}p{0.5cm}p{0.5cm}} 
        \toprule
        Functionalities & ShE & SE & DE\\
        \midrule
        Knowledge Graph &  $\times$ & $\times$  & \checkmark  \\
        Fused BERT Embeddings  & $\times$ & \checkmark & \checkmark   \\
        Weighted Average & \checkmark & $\times$ & \checkmark   \  \\
        Reinforcement Learning & $\times$ & $\times$ & \checkmark  \ \\
        \bottomrule
    \end{tabular}
    \caption{A comparison of ShE, SE, and DE Ensemble}
    \vspace{-2em}
    \label{table:functionality}
\end{table}

\section{EXPERIMENTS}
\textbf{Datasets:}  In our study, we employ nine benchmark classification datasets from the GLUE suite. The datasets are categorized in three categories of NLU 
- (a) Single Sentence (b) Inference (c) Similarity and Paraphrase. 
These include: (1) \textbf{CoLA} \cite{warstadt2019neural} for assessing grammatical correctness in English sentences, (2) \textbf{SST-2} \cite{socher2013recursive} for evaluating movie review sentiments, (3) \textbf{MRPC} \cite{madnani2012re} for determining whether two sentences are paraphrases, (4) \textbf{STS-B} \cite{conneau2018senteval} for rating the similarity between two sentences, modified in our study to binary labels, (5) \textbf{QQP} \cite{quora-question-pairs} for comparing similarity in pairs of Quora questions, (6) \textbf{MNLI} \cite{williams2018multi} addresses tasks involving pairs of sentences (Hypothesis and Premise) with labels: entailment, contradiction, and neutral. (7) \textbf{RTE} \cite{dagan2005pascal} aids in determining textual entailment within sentence pairs, (8) \textbf{QNLI} \cite{wang2018glue} involves a context sentence and a question to determine if the answer lies within the context, and (9) \textbf{WNLI} \cite{levesque2012winograd, wang2018glue} handles sentence coreference by discerning if an ambiguous pronoun refers to a designated target word. Dataset 1 and 2 consist of individual sentences, while Datasets 3-5 involve tasks related to measuring similarity and paraphrasing between a pair of sentences. Datasets 6-9 consists of natural language inferences tasks. The ensemble methods (i.e., ShE, SE, and DE) takes sentence $S_i$ as input. However datasets 3-7 comprises of a pair of sentences - $S_{i1}$ and $S_{i2}$. We process these sentences into a single input $S_i = S_{i1} \oplus S_{i2}$. 

\textbf{Metrics:} We assess our ensemble techniques using accuracy and the interrater reliability metric, Cohen's Kappa ($\kappa$). While accuracy is standard in GLUE tasks, $\kappa$ focuses more on reliability, which is a better measure to evaluate prediction uncertainty, specifically considering the chance behaviour of LMs \cite{mchugh2012interrater}. 
$\kappa$ is defined as $\kappa = \frac{p_o - p_e}{1 - p_e}$.  
In our study, we denote $p_e$ as the ground truth and $p_o$ as the class predicted by the model $\mathcal{M}$. Since $\kappa$ considers outcomes from two annotators, we consider the outcomes of $p_o$ and $p_e$ as our two annotators. According to McHugh \cite{mchugh2012interrater}, Interrater Reliability is directly proportional to $\kappa$. Consequently, $\kappa$ is directly proportional to the reliability of Language Models. This provides an answer to our first research question \textbf{RQ1:\textit{ Can we employ $\kappa$ to evaluate the reliability of LMs trained on GLUE
benchmarks?}}


\textbf{Experimental Setup:} We employ the BERT model to present our findings, as BERT is a streamlined model composed of a few million parameters, making it relatively simple and efficient. 
We consider the two variants of the BERT model i.e. $\text{BERT}_\text{{base}}$ and $\text{BERT}_\text{{large}}$ as our \textbf{\textit{baselines}}. We first compute $Z^{\text{BERT}_{\text{base}}}_{S_i}$ and $Z^{\text{BERT}_{\text{large}}}_{S_i}$ using equation \ref{eq:embeddings}, and then train using a NN classifier. 



\textbf{Reduced Embeddings:} To ensure an equitable comparison during the assessment of LMs, we employ Principal Component Analysis on $Z^{M_l}{S_i}$. The embedding dimensions of $BERT_\text{{base}}$ and $BERT_\text{{large}}$ are originally 768 and 1024, respectively, but we transform them into 100 dimensions each. ShE uses the condensed dimensions of $BERT_\text{{base}}$ and $BERT_\text{{large}}$. Initially, SE combines the embeddings of $BERT_\text{{base}}$ and $BERT_\text{{large}}$, resulting in an embedding dimension of 768+1024 = 1792, which is then further reduced to 100 dimensions. DE utilizes this fused embedding in conjunction with Wikipedia and ConceptNet embeddings. The embeddings from Wikipedia and ConceptNet are initially 500 and 300 dimensions, respectively, but are also reduced to 100 dimensions.

\textbf{Parameter Settings:} To ensure reproducibility, we partitioned the datasets using a random seed of 42. For ShE, SE, and DE, a NN was trained with a batch size of 8. Each model was tested on five different partitions (10\%, 15\%, 20\%, 25\%, and 30\%). The accuracy and $\kappa$ values presented in Table \ref{tab:results} represent the average performance across these partitions. We utilized the AdamW optimizer for batch normalization \cite{loshchilov2017decoupled}. To ensure a stable model-building process, we maintain a small learning rate of $2e^{-5}$ and weight decay of $1e^{-6}$.
\begin{table*}
    \begin{tabular}{p{1.5cm}|p{1.0cm}p{1.0cm}|p{1.0cm}p{0.5cm}|p{1.0cm}p{1.0cm}|p{1.0cm}|p{1.0cm}|p{1.0cm}p{1.0cm}}
        \toprule[1.5pt]
         \textbf{Dataset }& \multicolumn{2}{c}{$\mathbf{BERT_{base}}$} & \multicolumn{2}{|c}{\textbf{$\mathbf{BERT_{large}}  $}} & \multicolumn{2}{|c}{\textbf{ShE}} & \multicolumn{2}{|c}{\textbf{SE}} & \multicolumn{2}{|c}{\textbf{DE}} \\ 
         & \multicolumn{1}{c}{Accuracy} & \multicolumn{1}{c}{$\kappa$} &  \multicolumn{1}{|c}{Accuracy} & \multicolumn{1}{c}{$\kappa$} & \multicolumn{1}{|c}{Accuracy} & \multicolumn{1}{c}{$\kappa$} & 
         \multicolumn{1}{|c}{Accuracy} & \multicolumn{1}{c}{$\kappa$} & \multicolumn{1}|{c}{Accuracy} & \multicolumn{1}{c}{$\kappa$} 
         \\   
       \hline

        CoLA& \multicolumn{1}{c}{62.0} & \multicolumn{1}{c}{0.18} &  \multicolumn{1}{|c}{64.5} & \multicolumn{1}{c}{0.24} & \multicolumn{1}{|c}{67.34} & \multicolumn{1}{c}{0.28} & 
         \multicolumn{1}{|c}{\textbf{79.04}} & \multicolumn{1}{c}{\textbf{0.42}} & 
         \multicolumn{1}{|c}{\underline{72.88}} & \multicolumn{1}{c}{\underline{0.38}} 
         \\   
         
         MRPC& \multicolumn{1}{c}{56.0} & \multicolumn{1}{c}{0.11} &  \multicolumn{1}{|c}{52.3} & \multicolumn{1}{c}{0.02} & \multicolumn{1}{|c}{59.08} & \multicolumn{1}{c}{0.16} & 
         \multicolumn{1}{|c}{\textbf{73.08}} & \multicolumn{1}{c}{\textbf{0.35}} & 
         \multicolumn{1}{|c}{\underline{64.4}} & \multicolumn{1}{c}{\underline{0.28}}
         \\

          QNLI& \multicolumn{1}{c}{67.33} & \multicolumn{1}{c}{0.34} &  \multicolumn{1}{|c}{66.00} & \multicolumn{1}{c}{0.32} & \multicolumn{1}{|c}{\textbf{68.62}} & \multicolumn{1}{c}{\textbf{0.37}} & 
         \multicolumn{1}{|c}{67.34} & \multicolumn{1}{c}{0.35} & 
         \multicolumn{1}{|c}{\underline{67.65}} & \multicolumn{1}{c}{\underline{0.35}} 
         \\

          MNLI& \multicolumn{1}{c}{48.64} & \multicolumn{1}{c}{0.22} &  \multicolumn{1}{|c}{49.47} & \multicolumn{1}{c}{\textbf{0.25}} & \multicolumn{1}{|c}{\textbf{50.52}} & \multicolumn{1}{c}{\textbf{0.25}} & 
         \multicolumn{1}{|c}{49.9} & \multicolumn{1}{c}{\textbf{0.25}} & 
         \multicolumn{1}{|c}{\underline{50.0}} & \multicolumn{1}{c}{\textbf{0.25}} 
         \\

          QQP& \multicolumn{1}{c}{73.92} & \multicolumn{1}{c}{0.47} &  \multicolumn{1}{|c}{73.35} & \multicolumn{1}{c}{0.46} & \multicolumn{1}{|c}{74.80} & \multicolumn{1}{c}{0.49} & 
         \multicolumn{1}{|c}{75.12} & \multicolumn{1}{c}{0.50} & 
         \multicolumn{1}{|c}{\textbf{75.66}} & \multicolumn{1}{c}{\textbf{0.51}} 
         \\

          SST-2& \multicolumn{1}{c}{85.16} & \multicolumn{1}{c}{0.70} &  \multicolumn{1}{|c}{86.62} & \multicolumn{1}{c}{0.72} & \multicolumn{1}{|c}{87.5} & \multicolumn{1}{c}{0.74} & 
         \multicolumn{1}{|c}{87.9} & \multicolumn{1}{c}{\textbf{0.75}} & 
         \multicolumn{1}{|c}{\textbf{88.12}} & \multicolumn{1}{c}{\textbf{0.75}} 
         \\

          RTE& \multicolumn{1}{c}{52.45} & \multicolumn{1}{c}{0.04} &  \multicolumn{1}{|c}{48.86} & \multicolumn{1}{c}{0.00} & \multicolumn{1}{|c}{51.76} & \multicolumn{1}{c}{0.03} & 
         \multicolumn{1}{|c}{55.5} & \multicolumn{1}{c}{\textbf{0.12}} & 
         \multicolumn{1}{|c}{\textbf{56.03}} & \multicolumn{1}{c}{\textbf{0.12}} 
         \\

          STS-B& \multicolumn{1}{c}{63.01} & \multicolumn{1}{c}{0.25} &  \multicolumn{1}{|c}{62.31} & \multicolumn{1}{c}{0.24} & \multicolumn{1}{|c}{66.86} & \multicolumn{1}{c}{0.33} & 
         \multicolumn{1}{|c}{\textbf{76.6}} & \multicolumn{1}{c}{\textbf{0.52}} & 
         \multicolumn{1}{|c}{\underline{73.52}} & \multicolumn{1}{c}{\underline{0.47}} 
         \\

          WNLI& \multicolumn{1}{c}{49.93} & \multicolumn{1}{c}{0.006} &  \multicolumn{1}{|c}{51.72} & \multicolumn{1}{c}{0.03} & \multicolumn{1}{|c}{50.06} & \multicolumn{1}{c}{0.002} & 
         \multicolumn{1}{|c}{33.5} & \multicolumn{1}{c}{0.1} & 
         \multicolumn{1}{|c}{\textbf{57.07}} & \multicolumn{1}{c}{\textbf{0.14}} 
         \\   
         \bottomrule[1.5pt]

         GLUE Avg & \multicolumn{1}{c}{62.04} & \multicolumn{1}{c}{0.26} &  \multicolumn{1}{|c}{61.68} & \multicolumn{1}{c}{0.25} & \multicolumn{1}{|c}{64.06} & \multicolumn{1}{c}{0.29} & 
         \multicolumn{1}{|c}{66.44} & \multicolumn{1}{c}{0.37} & 
         \multicolumn{1}{|c}{\textbf{67.25}} & \multicolumn{1}{c}{0.36} 
         \\  
         
         \bottomrule[1.5pt]
    \end{tabular}
\caption{Performance Metrics - The accuracy and Cohen's Kappa ($\kappa$) for individual BERT models are compared with three variations of ensembles: ShE, SE - both without incorporating knowledge, and DE which includes knowledge. These models were assessed on the GLUE benchmark. The reported accuracy and $\kappa$ values are averages derived from 5 different split data, as elaborated in Section 3.3. In every instance, the ensembles outperformed the baseline models. Among them, DE has the \textbf{best} results in 4 tasks, and impressively attaining the \underline{second best performance} in the remaining 5 tasks.
}
    \label{tab:results}
\end{table*}

\section{RESULTS and DISCUSSION}
This section describes the experimental outcomes and addresses \textbf{RQ2}, and \textbf{RQ3}. \textbf{RQ1} has been addressed in Section 3 (Metrics).

\textbf{Overall Performance:} As outlined in Table \ref{tab:results}, we showcase accuracy and $\kappa$, of our ensemble techniques across nine GLUE datasets.
The results clearly indicate that:
(1) Our Ensemble Models uniformly outperform the baselines (individual BERT Models) across all nine GLUE tasks.
(2) ShE is the highest-performing model in two GLUE tasks.
(3) SE claims the lead in three GLUE tasks.
(4) DE excels in four GLUE tasks, the highest among the baselines and other ensembles (ShE and SE). DE's average accuracy surpasses $BERT_{base}$ by 5.21\% and $BERT_{large}$ by 5.57\% in the GLUE tasks, solidifying its position as the best-achieving model compared to the baselines (See Table \ref{tab:results}).


\textbf{RQ2:} \textbf{\textit{Considering the language models as annotators, is it possible to enhance $\kappa$ by strategically ensembling LMs? }}
Table \ref{tab:results} reveals that all ensemble models experience an increase in $\kappa$ score. It shows that SE achieves the highest $\kappa$, with DE also making significant strides. On average, there is a 0.12 increment in the $\kappa$ score compared to the baselines. Individual models often exhibit uncertainty in their predictions \cite{zhou2023navigating}. However, by strategically combining these models (i.e., ShE, SE, and DE), their weaknesses are counterbalanced by focusing on confident outcomes, elevating the $\kappa$ values.

\textbf{RQ3:} \textbf{\textit{Given that crowd workers frequently resort to external knowledge to augment the quality of annotations, can the infusion
of external knowledge during ensembling improve overall reliability?}}
DE incorporates knowledge from external sources. Table \ref{tab:results} demonstrates that DE achieves the best accuracy results compared to baselines and also sees an average increase in $\kappa$ score by 0.11, indicating that adding knowledge boosts the model's overall reliability. However, it’s noteworthy that SE records a marginally superior $\kappa$ score than DE. \textbf{\textit{This observation emphasizes that while increased accuracy often implies enhanced reliability, this isn't necessarily a universal truth}.}

\begin{figure}[htbp]
    \centering
    \begin{minipage}{0.48\textwidth}
        \centering
        \includegraphics[width=\linewidth]{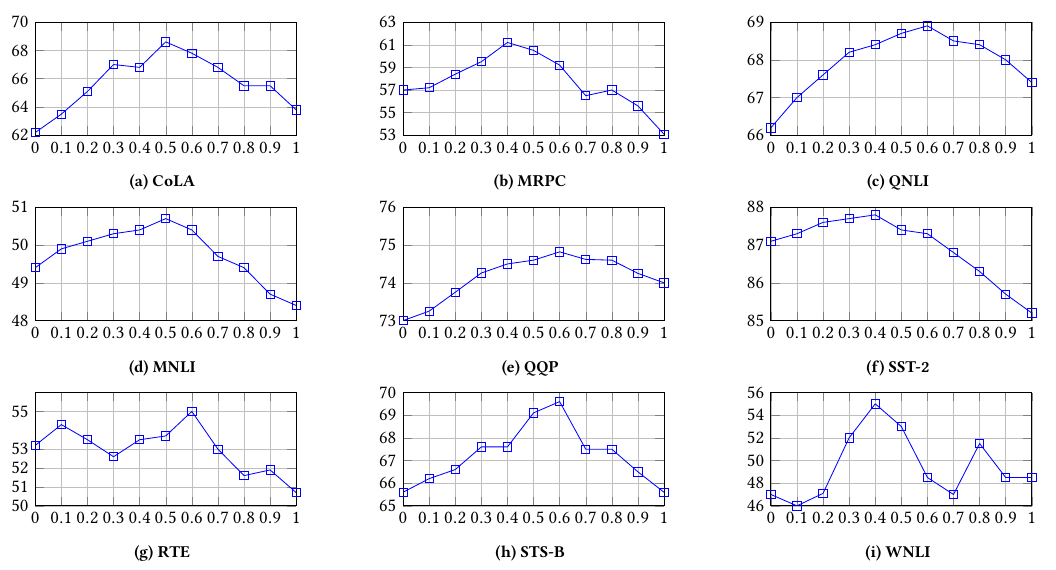}
        \caption{Ablation Study of ShE for GLUE datasets. x-axis represents $\bm{\alpha \in [0, 1]}$. The y-axis represents the accuracy. $\bm{\alpha = 0}$ denotes the performance of $\bm{\text{BERT}_{\text{large}}}$ and $\bm{\alpha = 1}$ denotes the performance of $\bm{\text{BERT}_{\text{base}}}$.}
        \label{fig:ablation_study}
    \end{minipage}
    \hfill
    \begin{minipage}{0.48\textwidth}
        \centering
        \includegraphics[width=\linewidth]{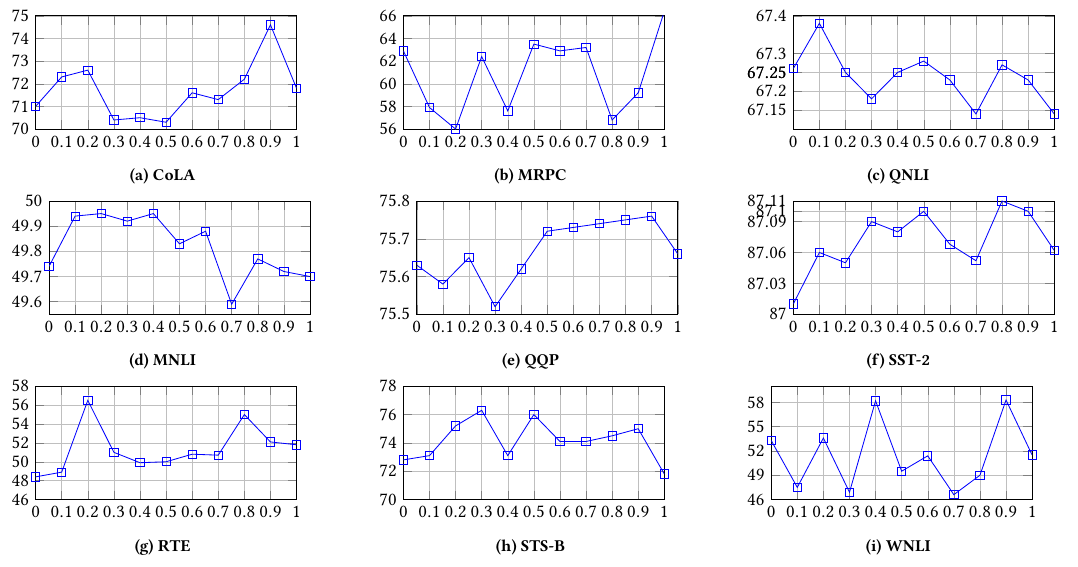}
        \caption{Ablation Study of DE for GLUE datasets. x-axis represents \bm{$\beta \in [0,1]$}. The y-axis represents the accuracy. $\bm{\alpha = 0}$ represents the knowledge infusion from CNet and $\bm{\alpha = 1}$ represents the knowledge infusion from Wiki. }
        \label{fig:ablation_study_de}
    \end{minipage}
\end{figure}

\textbf{Ablation Study:} This section entails ablation for the ensembles. \textbf{(1) ShE - } We compare $\alpha \in [0,1]$ as shown in Figure \ref{fig:ablation_study}. $\alpha = 1$ shows the performance of $\text{BERT}_{\text{base}}$, whereas $\alpha = 0$ represents $\text{BERT}_{\text{large}}$ as described in Equation \ref{eq:she1}. For single-sentence tasks, it is observed that the model performs best when there is an equal mixture ($\alpha \in [4,5]$) from both BERT variants 
In similarity tasks, for QQP and STS-B datasets, the model's performance is influenced by $\text{BERT}_\text{{base}}$ since $\alpha = 0.6$; whereas for MRPC it is more influenced by  $\text{BERT}_\text{{large}}$ as $\alpha = 0.4$. In inference tasks, for QNLI and RTE datasets, the model's performance is influenced by $\text{BERT}_\text{{base}}$ since $\alpha = 0.6$ whereas, for WNLI, it is influenced by  $\text{BERT}_\text{{large}}$ as  $\alpha = 0.4$. MNLI performs best when there is an equal contribution from both BERT variants. These results show that a model trained on lesser parameters ($\text{BERT}_{\text{base}}$) sometimes tends to perform better than a model trained on more parameters ($\text{BERT}_{\text{large}}$). 
\textbf{(2) SE - } No ablation study exists for SE because this ensemble consisted of a fusion of embeddings. \textbf{(3) DE - } Figure \ref{fig:ablation_study_de} displays the outcomes of DE for $\beta$.  $\beta$ regulates the extent of knowledge integration from the Knowledge Graphs (Section 2). From equation \ref{eq:sim_metric}, $\beta = 1$ denotes the knowledge infusion of Wiki whereas $\beta = 0$ considers the knowledge infusion from CNet. For single-sentence tasks, the model is highly influenced by adding knowledge from Wiki because the model gives the best performance at $\beta = 0.9$. there is an equal mixture ($\beta \in [4,5]$) from $\text{BERT}_{\text{base}}$ and $\text{BERT}_{\text{large}}$. In similarity tasks, it is found that for QQP and MRPC, the model's performance is highly influenced by Wiki because $\beta= 0.9$ and 1.0, respectively. In the case of STS-B, the model performs better when there is an equal mixture of both KGs. It performs equally well with CNet at $\beta = 0.3$. For Inference tasks, it can be seen that for all four datasets, the model's performance is highly influenced by CNet since $\beta \in [0.1,0.4]$. For WNLI, the model performs equally well with Wiki as $\beta =0.9$. The results for Single Sentence tasks and Similarity tasks showcase that adding Wiki is crucial in improving the model. 
However, its addition yields a contrasting outcome for Inference tasks, where CNet significantly enhances model improvement. This is because inference tasks rely on common sense knowledge, effectively captured by CNet \cite{wang2019improving}. 
\vspace{-0.5em}
\section{CONCLUSION AND FUTURE WORK}
In this research, we introduce ensembles of LMs to empirically assess their practicality, with particular emphasis on mitigating the inconsistent and unreliable nature of individual LMs.  All three BERT-ensembles showcase an enhancement in both accuracy and reliability over baselines. Additionally, combining LMs with a classifier whose loss is tuned by the RL method and integrating knowledge graphs contributes to significant accuracy improvement. Cohen's Kappa $\kappa$ was used to measure the LMs' reliability, showing that ensembling coupled with knowledge incorporation bolsters the LMs. However, it is important to note that improved accuracy doesn't necessarily translate to higher reliability. Future research avenues encompass the development of superior ensemble techniques and the evaluation of LMs using reliability metrics. We aim to examine our ensemble models on real-world datasets to check their reliability and performance in domain-specific applications.

\textbf{Acknowledgement:}
We sincerely thank Dr. Abhishek Kumar Umrawal for providing his valuable guidance throughout the research and Dr. Charles Nicholas for supporting the overhead costs involved with the research. 

\newpage
\bibliographystyle{ACM-Reference-Format}
\bibliography{sample-base}

\end{document}